\documentclass[11pt]{article}

\usepackage[utf8]{inputenc} 
\usepackage{times}
\usepackage[T1]{fontenc}    
\usepackage{hyperref}       
\usepackage{url}            
\usepackage{amsfonts}       
\usepackage{nicefrac}       
\usepackage{microtype}      
\usepackage{enumerate}
\usepackage{graphicx}

\usepackage{algorithm}
\usepackage[noend]{algpseudocode}

\newif\ifproofs
\proofstrue

\usepackage{natbib}
\bibpunct{(}{)}{;}{a}{,}{,}

\usepackage{aixi}
\usepackage{mythm}

\def\EP{\mathrm{EP}}

\begin{document}

\title{Exploration Potential}

\author{Jan Leike \\
    Future of Humanity Institute \\
    University of Oxford
}

\maketitle

\begin{abstract}
We introduce \emph{exploration potential},
a quantity that measures how much a reinforcement learning agent
has explored its environment class.
In contrast to information gain,
exploration potential takes the problem's reward structure into account.
This leads to an exploration criterion
that is both necessary and sufficient
for asymptotic optimality
(learning to act optimally across the entire environment class).
Our experiments in multi-armed bandits
use exploration potential to illustrate how different algorithms
make the tradeoff between exploration and exploitation.
\end{abstract}


\section{Introduction}
\label{sec:introduction}

Good exploration strategies are currently a major obstacle
for reinforcement learning (RL).
The state of the art in deep RL~\citep{MKSRV+:2015deepQ,MBMG+2016DQN}
relies on $\varepsilon$-greedy policies:
in every time step, the agent takes a random action
with some probability.
Yet $\varepsilon$-greedy is a poor exploration strategy
and for environments with sparse rewards it is quite ineffective
(for example the Atari game `Montezuma's Revenge'):
it just takes too long until the agent randomwalks into the first reward.

More sophisticated exploration strategies have been proposed:
using information gain about the environment%
~\citep{SGS:2011infogain,OLH:2013ksa,HCDSDA:2016explore}
or pseudo-count~\citep{BSOSSM:2016explore}.
In practice, these exploration strategies are employed by adding an exploration bonus (`intrinsic motivation')
to the reward signal~\citep{Schmidhuber:2010everything}.
While the methods above require the agent to have a model of its environment
and formalize the strategy `explore by going to where the model has high uncertainty,'
there are also model-free strategies like the
automatic discovery of options proposed by \citet{MB:2016options}.
However, none of these explicit exploration strategies
take the problem's reward structure into account.
Intuitively, we want to explore more in parts of the environment
where the reward is high and less where it is low.
This is readily exposed in optimistic policies like UCRL~\citep{JOA:2010UCRL}
and stochastic policies like PSRL~\citep{Strens:2000},
but these do not make the exploration/exploitation tradeoff explicitly.

In this paper, we propose \emph{exploration potential},
a quantity that measures reward-directed exploration.
We consider \emph{model-based} reinforcement learning
in partially or fully observable domains.
Informally, exploration potential is
the Bayes-expected absolute deviation of the value of optimal policies.
Exploration potential is
similar to information gain about the environment,
but explicitly takes the problem's reward structure into account.
We show that this
leads to a exploration criterion that
is both \emph{necessary} and \emph{sufficient}
for asymptotic optimality
(learning to act optimally across an environment class):
a reinforcement learning agent learns to act optimal in the limit
\emph{if and only if} the exploration potential converges to $0$.
As such, exploration potential captures the essence of
what it means to `explore the right amount'.

Another exploration quantity that is both necessary and sufficient
for asymptotic optimality is information gain about the optimal policy~\citep{Russo:2014,RCV:2016infomax}.
In contrast to exploration potential, it is not measured on the scale of rewards, making an explicit value-of-information tradeoff more difficult.

For example, consider a 3-armed Gaussian bandit problem with means
$0.6$, $0.5$, and $-1$.
The information content is identical in every arm.
Hence an exploration strategy based on maximizing information gain about the environment
would query the third arm, which is easily identifiable as suboptimal,
too frequently (linearly versus logarithmically).
This arm provides information,
but this information is not very useful for solving the reinforcement learning task.
In contrast, an exploration potential based exploration strategy
concentrates its exploration on the first two arms.

\section{Preliminaries and Notation}
\label{sec:preliminaries-and-notation}

A reinforcement learning agent interacts with an environment in cycles:
at time step $t$ the agent chooses an \emph{action} $a_t$ and
receives a \emph{percept} $e_t = (o_t, r_t)$
consisting of an \emph{observation} $o_t$
and a \emph{reward} $r_t \in [0, 1]$;
the cycle then repeats for $t + 1$.
We use $\ae_{<t}$ to denote a history of length $t - 1$.
With abuse of notation,
we treat histories both as outcomes and as random variables.

A \emph{policy} is a function
mapping a history $\ae_{<t}$ and an action $a$ to
the probability $\pi(a \mid \ae_{<t})$ of taking action $a$ after seeing history $\ae_{<t}$.
An \emph{environment} is a function
mapping a history $\ae_{1:t}$ to
the probability $\nu(e_t \mid \ae_{<t} a_t)$
of generating percept $e_t$ after this history $\ae_{<t} a_t$.
A policy $\pi$ and an environment $\nu$ generate
a probability measure $\nu^\pi$ over infinite histories,
the expectation over this measure is denoted with $\EE^\pi_\nu$.
The \emph{value} of a policy $\pi$ in an environment $\nu$
given history $\ae_{<t}$ is defined as
\[
   V_\nu^\pi(\ae_{<t})
:= (1 - \gamma)
     \EE_\nu^\pi\left[ \sum_{k=t}^\infty \gamma^k r_k \,\middle|\, \ae_{<t} \right],
\]
where $\gamma \in (0, 1)$ is the discount factor.
The \emph{optimal value} is defined as
$V^*_\nu(\ae_{<t}) := \sup_\pi V^\pi_\nu(\ae_{<t})$, and the \emph{optimal policy} is
$\pi^*_\nu := \argmax_\pi V^\pi_\nu$.
We use $\mu$ to denote the true environment.

We assume the nonparametric setting:
let $\M$ denote a countable class of environments
containing the true environment $\mu$.
Let $w \in \Delta\M$ be a prior probability distribution on $\M$. After observing the history $\ae_{<t}$
the prior $w$ is updated to the posterior
$w(\nu \mid \ae_{<t}) := w(\nu) \nu(\ae_{<t}) / (\sum_{\rho \in \M} w(\rho) \rho(\ae_{<t}))$.
A policy $\pi$ is \emph{asymptotically optimal in mean} iff
for every $\mu \in \M$,
$\EE_\mu^\pi[ V^*_\mu(\ae_{<t}) - V^\pi_\mu(\ae_{<t})] \to 0$ as $t \to \infty$.

\section{Exploration Potential}
\label{sec:exploration-potential}

We consider \emph{model-based} reinforcement learning
where the agent learns a model of its environment.
With this model, we can estimate the value of any candidate policy.
Concretely, let $\hat{V}_t^\pi$ denote our estimate
of the value of the policy $\pi$ at time step $t$.
We assume that the agent's learning algorithm satisfies
\emph{on-policy value convergence}~(OPVC):
\begin{equation}\label{eq:on-policy-value-convergence}
V^\pi_\mu(\ae_{<t}) - \hat{V}_t^\pi(\ae_{<t}) \to 0
\text{ as $t \to \infty$ $\mu^\pi$-almost surely.}
\end{equation}
This does \emph{not} imply that our model of the environment converges to the truth,
only that we learn to predict the value of the policy that we are following.
On-policy value convergence does not require that we learn to predict off-policy,
i.e., the value of other policies.
In particular, we might not learn to predict the value of the $\mu$-optimal policy $\pi^*_\mu$.

For example, a Bayesian mixture or an MDL-based estimator
both satisfy OPVC if the true environment is the environment class;
for more details, see \citet[Sec.~4.2.3]{Leike:2016}.

We define the $\hat{V}_t$-greedy policy as
$\pi^*_{\hat{V}} := \argmax_\pi \hat{V}^\pi_t$.

\subsection{Definition}
\label{sec:exploration-potential-def}

\begin{definition}[Exploration Potential]
\label{def:exploration-potential}
Let $\M$ be a class of environments and let $\ae_{<t}$ be a history.
The \emph{exploration potential} is defined as
\[
   \EP_\M(\ae_{<t})
:= \sum_{\nu \in \M} w(\nu \mid \ae_{<t})
     \left| V_\nu^{\pi^*_\nu}(\ae_{<t}) - \hat{V}_t^{\pi^*_\nu}(\ae_{<t}) \right|.
\]
\end{definition}

Intuitively, $\EP$ captures the amount of exploration
that is still required before having learned the entire environment class.
Asymptotically the posterior concentrates around
environments that are compatible with the current environment.
EP then quantifies how well the model $\hat{V}_t$ understands
the value of the compatible environments' optimal policies.

\begin{remark}[Properties of $\EP$]
\label{rem:properties-of-EP}
~
\begin{enumerate}[(i)]
\item $\EP_\M$ depends neither on the true environment $\mu$,
    nor on the agent's policy $\pi$.
\item $\EP_\M$ depends on the choice of the prior $w$ and
    on the agent's model of the world $\hat{V}_t$.
\item
$
     0
\leq \EP_\M(\ae_{<t})
\leq 1
$
for all histories $\ae_{<t}$.
\end{enumerate}
\end{remark}

The last item follows from the fact that
the posterior $w(\,\cdot \mid \ae_{<t})$ and the value function $V$
are bounded between $0$ and $1$.

\subsection{Sufficiency}
\label{ssec:sufficiency}

\begin{proposition}[Bound on Optimality]
\label{prop:full-exploration}
For all $\mu \in \M$,
\[
       V^*_\mu(\ae_{<t}) - V^{\pi^*_{\hat{V}}}_\mu(\ae_{<t})
~\leq~ \hat{V}_t^*(\ae_{<t}) - V^{\pi^*_{\hat{V}}}_\mu(\ae_{<t})
       ~+~ \frac{\EP_\M(\ae_{<t})}{w(\mu \mid \ae_{<t})}.
\]
\end{proposition}
\ifproofs
\begin{proof}
\begin{align*}
     \left| V^*_\mu - \hat{V}_t^{\pi^*_\mu} \right|
=    \frac{w(\mu \mid \ae_{<t})}{w(\mu \mid \ae_{<t})}
       \left| V^*_\mu - \hat{V}_t^{\pi^*_\mu} \right|
\leq \sum_{\nu \in \M} \frac{w(\nu \mid \ae_{<t})}{w(\mu \mid \ae_{<t})}
       \left| V^*_\nu - \hat{V}_t^{\pi^*_\nu} \right|
=    \frac{\EP_\M}{w(\mu \mid \ae_{<t})}
\end{align*}
Therefore
\[
  V^*_\mu - V^{\pi^*_{\hat{V}}}_\mu
~~~= \underbrace{V^*_\mu - \hat{V}_t^{\pi^*_\mu}}_{\leq \EP(\ae_{<t}) / w(\mu \mid \ae_{<t})}
  +~~~ \underbrace{\hat{V}_t^{\pi^*_\mu} - \hat{V}_t^*}_{\leq 0}
  ~~+~~~ \hat{V}^*_t - V^{\pi^*_{\hat{V}}}_\mu.
\qedhere
\]
\end{proof}
\fi

The bound of \autoref{prop:full-exploration} is to be understood as follows.
\[
\underbrace{V^*_\mu(\ae_{<t}) - V^{\pi^*_{\hat{V}}}_\mu(\ae_{<t})}_{\text{optimality of the greedy policy}}
~~\leq~~
\underbrace{\hat{V}^*_t(\ae_{<t}) - V^{\pi^*_{\hat{V}}}_\mu(\ae_{<t})}_{\text{OPVC}}
~~~+ \underbrace{\EP(\ae_{<t})}_{\text{exploration potential}}
/ \underbrace{w(\mu \mid \ae_{<t})}_{\text{posterior}}
\]
If we switch to the greedy policy $\pi^*_{\hat{V}}$,
then $\hat{V}_t^* - V^{\pi^*_{\hat{V}}}_\mu \to 0$
due to on-policy value convergence~\eqref{eq:on-policy-value-convergence}.
This reflects how well
the agent learned the environment's response to the Bayes-optimal policy.
Generally, following the greedy policy does not yield
enough exploration for $\EP$ to converge to $0$.
In order to get a policy $\pi$ that is asymptotically optimal,
we have to combine an exploration policy which ensures that $\EP \to 0$
and then gradually phase out exploration by switching to
the $\pi^*_{\hat{V}}$-greedy policy.
Because of property (i),
the agent can compute its current $\EP$ value
and thus check how close it is to $0$.
The higher the prior belief in the true environment $\mu$,
the smaller this value will be (in expectation).

\subsection{Necessity}
\label{sec:necessity}

\begin{definition}[Policy Convergence]
\label{def:policy-convergence}
Let $\pi$ and $\pi'$ be two policies.
We say the policy $\pi$ \emph{converges to $\pi'$ in $\mu^\pi$-probability} iff
$|\hat{V}^\pi_t(\ae_{<t}) - \hat{V}^{\pi'}_t(\ae_{<t})| \to 0$ as $t \to \infty$
in $\hat{V}$.
\end{definition}

We assume that $\hat{V}_t$ is continuous in the policy argument.
If $\pi$ converges to $\pi'$ in total variation
in the sense that
$\pi(a \mid \ae_{<k}) - \pi'(a \mid \ae_{<k}) \to 0$ for all actions $a$
and $k \geq t$,
then $\pi$ converges to $\pi'$ in $\hat{V}$.

\begin{definition}[Strongly Unique Optimal Policy]
\label{def:strongly-unique-optimal-policy}
An environment $\mu$ admits a \emph{strongly unique optimal policy} iff
there is a $\mu$-optimal policy $\pi^*_\mu$ such that for all policies $\pi$
if
\[
V^*_\mu(\ae_{<t}) - V^\pi_\mu(\ae_{<t}) \to 0
\text{ in $\mu^\pi$-probability,}
\]
then $\pi$ converges to $\pi^*_\mu$ in $\hat{V}$.
\end{definition}

Assuming that $\hat{V}_t^\pi$ is continuous is $\pi$,
an environment $\mu$ has a unique optimal policy if there are no ties
in $\argmax_a V^*_\mu(\ae_{<t}a)$.
Admitting a strongly unique optimal policy is an even stronger requirement
because it requires that there exist no other policies that
approach the optimal value asymptotically but take different actions
(i.e., there is a constant gap in the argmax).
For any finite-state (PO)MDP
with a unique optimal policy that policy is also strongly unique.

\begin{proposition}[Asymptotic Optimality $\Rightarrow \EP \to 0$]
\label{prop:insufficient-exploration}
If the policy $\pi$ is asymptotically optimal in mean in the environment class $\M$
and each environment $\nu \in \M$ admits a strongly unique optimal policy,
then $\EP_\M \to 0$ in $\mu^\pi$-probability for all $\mu \in \M$.
\end{proposition}
\ifproofs
\begin{proof}
Since $\pi$ is asymptotically optimal in mean in $\M$,
we have that $V^*_\mu - V^\pi_\mu \to 0$
and since $\mu$ admits a strongly unique optimal policy,
$\pi$ converges to $\pi^*_\mu$ in $\mu^\pi$-probability,
thus $\hat{V}_t^\pi - \hat{V}_t^{\pi^*_\mu} \to 0$.
By on-policy value convergence $V^\pi_\mu - \hat{V}_t^\pi \to 0$.
Therefore
\[
    V^*_\mu - \hat{V}_t^{\pi^*_\mu}
=   V^*_\mu - V^\pi_\mu + V^\pi_\mu - \hat{V}_t^\pi + \hat{V}_t^\pi - \hat{V}_t^{\pi^*_\mu}
\to 0
\]
and thus
\begin{equation}\label{eq:conv}
\EE^\pi_\mu \left|
    V_\mu^{\pi^*_\mu}(\ae_{<t}) - \hat{V}_t^{\pi^*_\mu}(\ae_{<t})
\right| \to 0
\text{ for all $\mu \in \M$.}
\end{equation}
Now
\begin{align*}
      \EE^\pi_\mu[ \EP_\M(\ae_{<t}) ]
&=    \EE^\pi_\mu \left[
        \sum_{\nu \in \M} w(\nu \mid \ae_{<t})
          \left| V_\nu^{\pi^*_\nu}(\ae_{<t}) - \hat{V}_t^{\pi^*_\nu}(\ae_{<t}) \right|
      \right] \\
&\leq \frac{1}{w(\mu)} \EE^\pi_\xi \left[
        \sum_{\nu \in \M} w(\nu \mid \ae_{<t})
          \left| V_\nu^{\pi^*_\nu}(\ae_{<t}) - \hat{V}_t^{\pi^*_\nu}(\ae_{<t}) \right|
      \right] \\
&=    \frac{1}{w(\mu)} \sum_{\nu \in \M} w(\nu) \EE^\pi_\xi \left[
        \frac{\nu^\pi(\ae_{<t})}{\xi^\pi(\ae_{<t})}
          \left| V_\nu^{\pi^*_\nu}(\ae_{<t}) - \hat{V}_t^{\pi^*_\nu}(\ae_{<t}) \right|
      \right] \\
&=    \frac{1}{w(\mu)} \sum_{\nu \in \M} w(\nu) \EE^\pi_\nu
          \left| V_\nu^{\pi^*_\nu}(\ae_{<t}) - \hat{V}_t^{\pi^*_\nu}(\ae_{<t}) \right|
 \to  0
\end{align*}
by \eqref{eq:conv} and \citet[Lem.~5.28ii]{Hutter:2005}.
\end{proof}
\fi

If we don't require the condition on strongly unique optimal policies,
then the policy $\pi$ could be asymptotically optimal
while $\EP \not\to 0$:
there might be another policy $\pi'$
that is very different from any optimal policy $\pi^*_\mu$,
but whose $\mu$-value approaches the optimal value:
$V^*_\mu - V^{\pi'}_\mu \to 0$ as $t \to \infty$.
Our policy $\pi$ could converge to $\pi'$ without $\EP$ converging to $0$.

\section{Exploration Potential in Multi-Armed Bandits}
\label{sec:bandits}

\begin{figure}[t]
\begin{center}
\includegraphics[width=\textwidth]{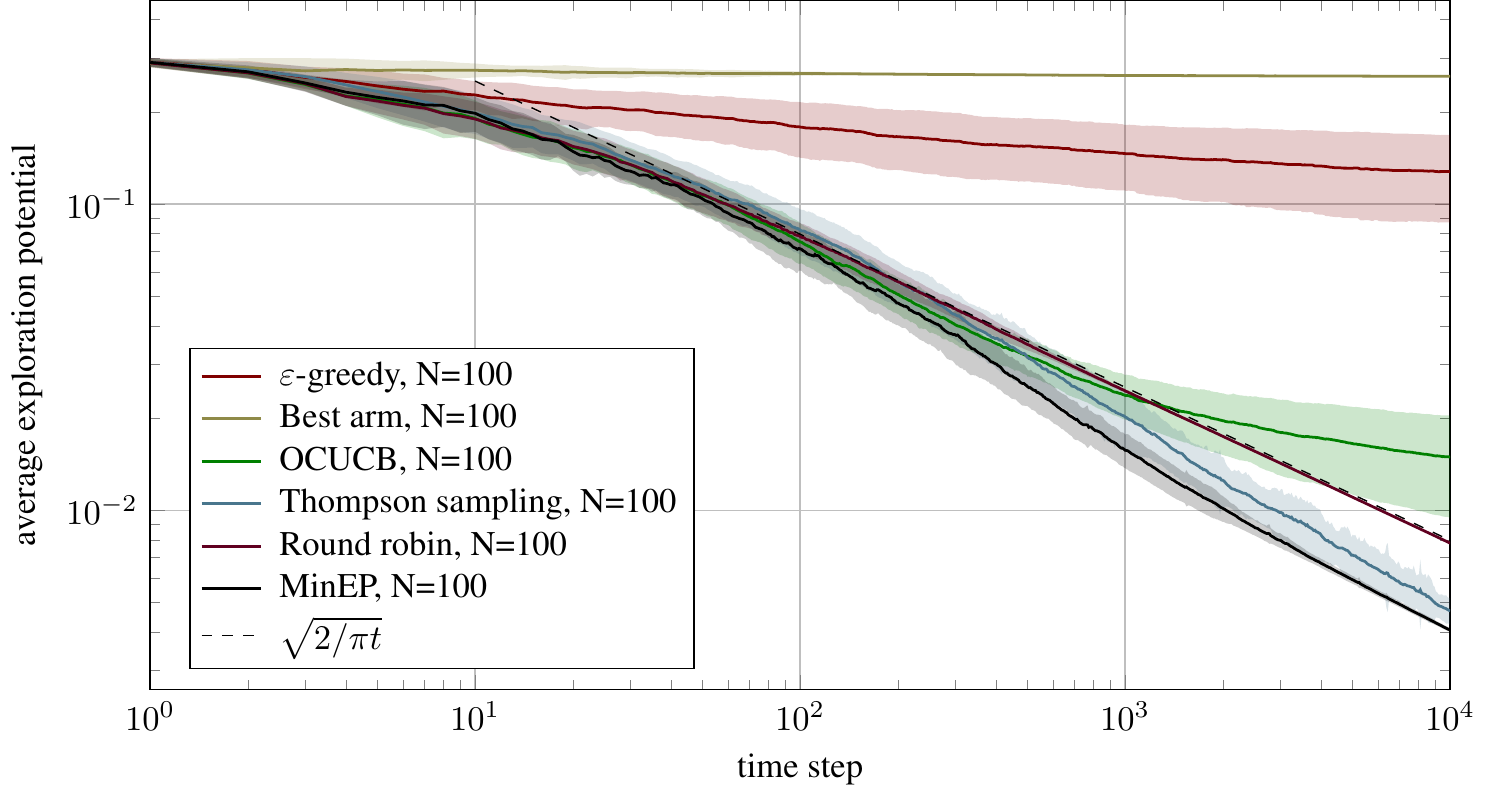}
\end{center}
\caption{
Exploration potential over time for different bandit algorithms
in the Bernoulli bandit with arms $0.6, 0.5, 0.4, 0.4$%
~(double logarithmic plot);
shaded regions correspond to one standard deviation.
Lower exploration potential means more exploration.
The notable change in slope in around time steps 20--80
stems from the fact that it takes about that long to
reliably distinguish the first two arms.
The dashed line corresponds to the optimal asymptotic rate
of $t^{-1/2}$.
}\label{fig:exploration-potential}
\end{figure}

In this section we use experiments with
multi-armed Bernoulli bandits
to illustrate the properties of exploration potential.
The class of Bernoulli bandits is $\Theta = [0, 1]^k$
(the arms' means).
In each time step, the agent chooses an action (arm)
$i \in \{ 1, \ldots, k \}$
and receives a reward $r_t \sim Bernoulli(\theta_i^*)$
where $\theta^* \in \Theta$ is the true environment.
Since $\Theta$ is uncountable,
exploration potential is defined with an integral instead of a sum:
\[
   \EP_\Theta(\ae_{<t})
:= \int_\Theta p( \theta \mid \ae_{<t}) | \theta_{j(\theta)} - \hat\theta_{j(\theta)} | d\theta
\]
where $p(\theta \mid \ae_{<t})$ is
the posterior distribution given the history $\ae_{<t}$,
$\hat\theta := \int_\Theta \theta p(\theta \mid \ae_{<t})d\theta$ is
the Bayes-mean parameter, and
$j(\theta) := \argmax_i \theta_i$ is
the index of the best arm accoding to $\theta$.

\autoref{fig:exploration-potential} shows the exploration potential
of several bandit algorithms,
illustrating how much each algorithm explores.
Notably, optimally confident UCB~\citep{Lattimore:2015OCUCB}
stops exploring around time step 700 and focuses on exploitation
(because in contrast to the other algorithms it knows the horizon).
Thompson sampling, round robin (alternate between all arms),
and $\varepsilon$-greedy
explore continuously (but $\varepsilon$-greedy is less effective).
The optimal strategy (always pull the first arm) never explores and hence its exploration potential decreases only slightly.

\begin{figure}[t]
\begin{center}
\includegraphics[width=\textwidth]{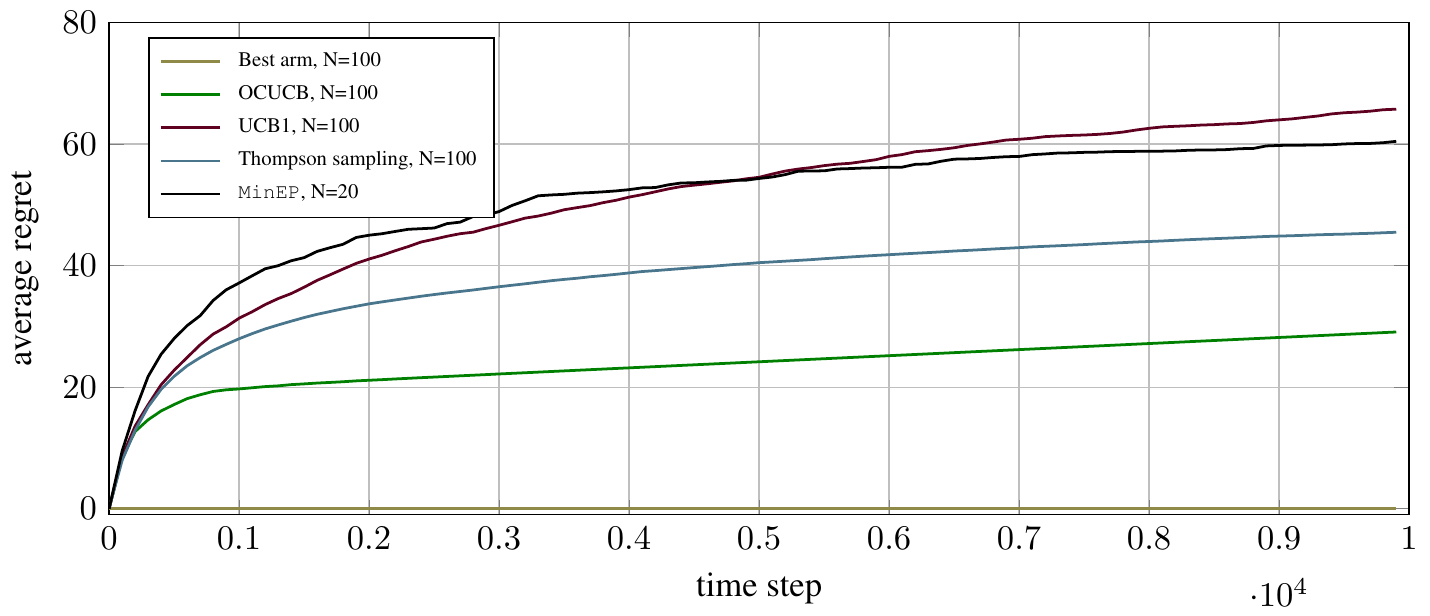}
\end{center}
\caption{
Average regret over time for different bandit algorithms
in the Bernoulli bandit with arms $0.6, 0.5, 0.4, 0.4$.
\texttt{MinEP} outperforms UCB1~\citep{Auer:2002UCB} after 10\,000 steps, but neither Thompson sampling nor OCUCB.
}\label{fig:regret}
\end{figure}

Exploration potential naturally gives rise to an exploration strategy: greedily minimize Bayes-expected exploration potential
(\texttt{MinEP}); see \autoref{alg:minEP}.
This strategy unsurprisingly explores more than all the other algorithms when measured on exploration potential,
but in bandits it also turns out to be a decent exploitation strategy because it focuses its attention on the most promising arms.
For empirical performance see \autoref{fig:regret}.
However, \texttt{MinEP} is generally not a good exploitation strategy
in more complex environments like MDPs.

\begin{algorithm}
\begin{center}
\begin{algorithmic}[1]
\For{$t \in \mathbb{N}$}
	\State $a_t := \argmin_{a \in \A} \mathbb{E}_{e_t \sim \text{posterior}}[\EP(\ae_{<t} a e_t)]$
	\State take action $a_t$
	\State observe percept $e_t$
\EndFor
\end{algorithmic}
\end{center}
\caption{The \texttt{MinEP} Algorithm}
\label{alg:minEP}
\end{algorithm}

\section{Discussion}
\label{sec:discussion}

Several variants on the definition exploration potential given in \autoref{def:exploration-potential} are conceivable.
However, often they do not satisfy at least one of the properties
that make our definition appealing.
Either they break the necessity~(\autoref{prop:full-exploration}),
sufficiency~(\autoref{prop:insufficient-exploration}),
our proofs thereof,
or they make $\EP$ hard to compute.
For example, we could replace $|V^*_\nu - \hat{V}^{\pi^*_\nu}_t|$ by $|V^*_\nu - V^\pi_\nu|$ where $\pi$ is the agent's future policy. This preserves both necessesity and sufficiency, but relies on computing the agent's future policy. If the agent uses exploration potential for taking actions (e.g., for targeted exploration),
then this definition becomes a self-referential equation and might be very hard to solve.
Following \citet{Dearden+:1999}, we could consider $|V^*_\nu - \hat{V}^*_t|$ which has the convenient side-effect that it is model-free and therefore applies to more reinforcement learning algorithms.
However, in this case the necessity guarantee (\autoref{prop:insufficient-exploration}) requires the additional condition that the agent's policy converges to the greedy policy $\pi^*_{\hat{V}}$.
Moreover, this does not remove the dependence on a model since
we still need a model class $\M$ and a posterior.

Based on the recent successes in approximating information gain~\citep{HCDSDA:2016explore},
we are hopeful that exploration potential can also be approximated in practice.
Since computing the posterior is too costly for complex reinforcement learning problems,
we could (randomly) generate a few environments and
estimate the sum in \autoref{def:exploration-potential} with them.

In this paper we only scratch the surface on exploration potential
and leave many open questions.
Is this the correct definition?
What are good rates at which EP should converge to 0?
Is minimizing EP the most efficient exploration strategy?
Can we compute EP more efficiently than information gain?

\subsubsection*{Acknowledgments}
We thank Tor Lattimore, Marcus Hutter,
and our coworkers at the FHI for valuable feedback and discussion.


\bibliography{ai}

\end{document}